\pdfoutput=1

\documentclass[11pt]{article}

\usepackage{emnlp2021}

\usepackage{times}
\usepackage{latexsym}

\usepackage[T1]{fontenc}

\usepackage[utf8]{inputenc}

\usepackage{microtype}

\usepackage{graphicx}

\def\Funtional{DefiNNet}
\def\DefBERTplain{DefBERT}
\def\DefBERT{DefBERT$_{[CLS]}$}
\def\HeadBERT{DefBERT$_{Head}$}
\def\ExampleBERT{BERT$_{Head-Example}$}
\def\OOVBERT{BERT$_{word pieces}$}
\usepackage{multirow}

\usepackage{comment}
\usepackage{graphicx}
\usepackage{qtree}

\title{Lacking the embedding of a word? Look it up into a traditional dictionary}

\author{
Elena Sofia Ruzzetti$^{(*)}$  \\ {\bf Leonardo Ranaldi}$^{(\ddag,*)}$ \\ {\bf Michele Mastromattei}$^{(*)}$ \\
  $^{(*)}$ ART Group \\
  Department of Enterprise Engineering \\
  University of Rome Tor Vergata \\
  Viale del Politecnico, 1, 
  00133 Rome, Italy\\
  {\tt \small fabio.massimo.zanzotto@uniroma2.it} \\\And
 {Francesca Fallucchi}$^{(\ddag)}$ \\ {\bf Fabio Massimo Zanzotto}$^{(*)}$  \\
  $^{(\ddag)}$ Department of Innovation\\and Information Engineering\\
  Guglielmo Marconi University\\ 
Via Plinio 44, 
00193 Rome, Italy \\
  {\tt \small f.fallucchi@unimarconi.it} \\}

\begin{document}
\maketitle
\begin{abstract}
Word embeddings are powerful dictionaries, which may easily capture language variations. However, these dictionaries fail to give sense to rare words, which are surprisingly often covered by traditional dictionaries. 
In this paper, we propose to use definitions retrieved in traditional dictionaries to produce word embeddings for rare words. For this purpose, we introduce two methods: Definition Neural Network (DefiNNet) and Define BERT (DefBERT). In our experiments, DefiNNet and DefBERT significantly outperform state-of-the-art as well as baseline methods devised for producing embeddings of unknown words. In fact, DefiNNet significantly outperforms FastText, which implements a  method for the same task-based on n-grams, and DefBERT significantly outperforms the BERT method for OOV words. Then, definitions in traditional dictionaries are useful to build word embeddings for rare words.


\end{abstract}

\section{Introduction}

\begin{figure*}
    \centering
    \includegraphics[width=15cm]{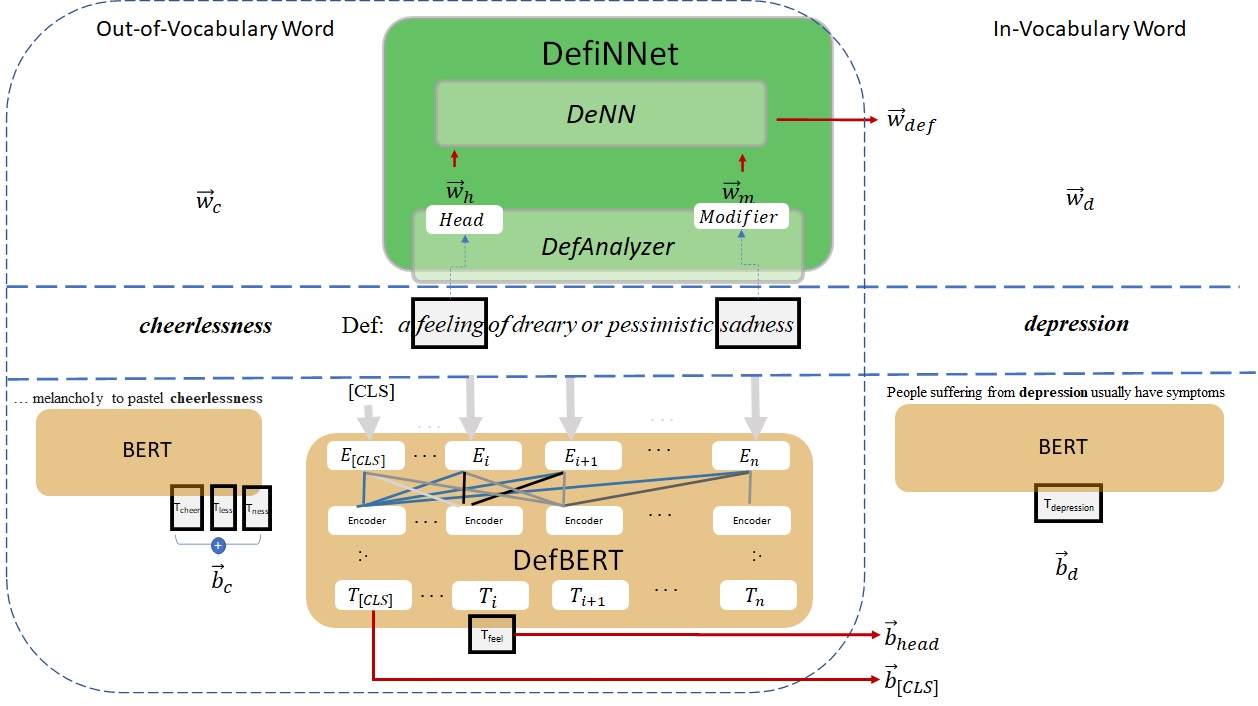}
    \caption{Exploiting definitions for Out-of-Vocabulary words: the \Funtional{} and the \DefBERTplain{} models. 
    }
    \label{fig:general_idea}
\end{figure*}

Words without meaning are like compasses without needle: pointless. Indeed, meaningless words lead compositionally to meaningless sentences and, consequently, to meaningless texts and conversations. 
Second language learners may grasp grammatical structures of sentences but, if they are unaware of meaning of single words in these sentences, they may fail to understand the whole sentences.  
This is the reason why large body of natural language processing research is devoted to devising ways to capture word meaning.

As language is a living body, distributional methods \cite{PantelTurney, mikolov2013efficient, pennington-etal-2014-glove}  are seen as the panacea to capture word meaning as opposed to more static models based on dictionaries \cite{wordnet-resource}.
Distributional methods may easily capture new meaning of existing words and, eventually, can easily assign meaning to emerging words. In fact, the different methods can scan corpora and derive the meaning of these new words by observing them in context \cite{Harris,Firth,Wittgenstein1953-WITPI-4}. Words are then represented as vectors -- now called word embeddings -- which are then used to feed neural networks to produce meaning for sentences \cite{10.5555/944919.944966,irsoy-cardie-2014-opinion, kalchbrenner-etal-2014-convolutional, tai-etal-2015-improved} and meaning for whole texts \cite{joulin-etal-2017-bag, 10.5555/2886521.2886636}.  

Distributional methods have a strong limitation: word meaning can be assigned only for words where sufficient contexts can be gathered. Rare words are not covered and become the classical out-of-vocabulary words, which may hinder the understanding of specific yet important sentences. To overcome this problem, n-grams based distributional models have emerged \cite{joulin2016bag} where word meaning is obtained by composing \emph{``meaning''} of n-grams forming a word.
These n-grams act as proto-morphemes and, hence, meaning of unknown words can be obtained by composing meaning of proto-morphemes derived for existing words. These proto-morphemes are the building blocks of word meaning. 


Traditional dictionaries can offer a solution to find meaning of rare words. 
They have been put aside since they cannot easily adapt to language evolution and they cannot easily provide distributed representations for 
neural networks. 

In this paper, we propose to use definitions in dictionaries to compositionally produce distributional representations for Out-Of-Vocabulary (OOV) words. Definitions in dictionaries are intended to describe the meaning of a word to a human reader. Then, we propose two models to exploit definitions to derive the meaning of OOV words: (1) Definition Neural Network (\Funtional{}), a simple neural network; (2) \DefBERTplain{}, a model based on pre-trained BERT.
We experimented with different tests and datasets derived from WordNet. Firstly, we determined if \Funtional{} and \DefBERTplain{} can learn a neural network to derive word meaning from definitions. Secondly, we aimed to establish whether \Funtional{} and \DefBERTplain{} can cover OOV words, which are not covered by word2vec \cite{mikolov2013efficient} or by the BERT pre-trained encoder, respectively. 
In our experiments, DefiNNet and DefBERT significantly outperform stete-of-the-art as well as baseline methods devised for producing embeddings of unknown words. In fact, DefiNNet significantly outperforms FastText \cite{joulin2016bag}, which implements a  method for the same task-based on n-grams, and DefBERT significantly outperforms the BERT method for OOV words. Then, definitions in traditional dictionaries are useful to build word embeddings for rare words.

\section{Background and Related Work}

Out-of-Vocabulary (OOV) words have been often a problem as these OOV words may hinder the applicability of many NLP systems. For example, if words are not included in a lexicon of a Probabilistic Context-Free Grammar, interpretations for sentences containing these words may have a null probability. Hence, solutions to this problem date back in time.

Recently, in the context of word embeddings, the most common solution is to use word n-grams \cite{joulin2016bag} or word pieces of variable length \cite{wu2016google} as proxies to model morphemes.
Embeddings are learned for 3-grams as well as for word pieces. In \citet{joulin2016bag} these 3-grams are then combined to obtain the embedding for the entire word. For example, the word \emph{cheerlessness}, which contains 3 morphemes (\emph{cheer}, \emph{less} and \emph{ness}), is modeled by using embeddings for $\vec{che}$, $\vec{hee}$, ..., $\vec{ess}$ in the 3-gram approach and by using embeddings for $\vec{cheer}$ and $\vec{lessness}$ in the word pieces approach. These embeddings are possibly capturing information about the related morphemes. In this way, OOV word embeddings are correlated with meaningful bits of observed words.
These models are clearly our baselines. 

In the study of OOV words for word embeddings, deriving word embeddings from dictionary definitions is, at the best of our knowledge, a novel approach. 
Dictionary definitions have been used in early attempts to train rudimentary compositional distributional semantic models 
\cite{zanzotto-etal-2010-estimating}, which aimed to build embeddings for sequences of two words. 

Universal sentence embedders (USEs) \cite{Conneau2018} can play an important role in this novel approach. In fact, definitions are particular sentences aiming to describe meaning of words. Therefore, USEs should obtain an embedding representing the meaning of a word by composing embeddings of words in the definition.


Moreover, deriving word embeddings from definitions can be seen as a semantic stress test of universal sentence embedders. Generally, the ability of USEs \cite{devlin-etal-2019-bert,yang2020xlnet,clark2020electra} to semantically model sentences is tested with end-to-end downstream tasks, for example, natural language inference (NLI) \cite{BERT-NLI-jiang-de-marneffe-2019-evaluating,raffel2020exploring,BERT-NLI-he2021deberta}, question-answering \cite{Zhang2019BERTFQ} as well as dialog systems \cite{wu2020todbert}. USEs such as BERT \cite{devlin-etal-2019-bert} are encoding semantic features in hidden layers \cite{Jawahar2019,miaschi2020linguistic}. This explains why these USEs are good at modeling semantics of sentences in downstream tasks. However, USEs' success in downstream tasks may be due to superficial heuristics (as supposed in \cite{mccoy2019right} for the NLI) and not to a deep modeling of semantic features.      
Therefore, our study can contribute to this debate. In fact, at the best of our knowledge, it is the first study aiming to investigate if USEs can model meaning by producing embedding for words starting from their definitions.

\section{Model}
\label{sec:model}
This section introduces our proposals to use definitions in generating embeddings for Out-of-Vocabulary words: Definition Neural Network (\Funtional{}) and BERT for Definitions (\DefBERTplain{}). Section \ref{sec:Notation} describe the  basic idea. Section \ref{sec:DefNN} describes the definition of the feed-forward neural network \Funtional{}. Finally, Section \ref{sec:DefBERT} describes how we used the Universal Sentence Embedder BERT in producing embeddings for definitions.

\subsection{Basic Idea}
\label{sec:Notation}

Our model stems from an observation: when someone step into an rare unknown word while reading, definitions in traditional dictionaries are the natural resource used to understand the meaning of this rare, out-of-one's-personal-dictionary word. Then, as people rely on dictionaries in order to understand meanings for unknown words, learners of word embeddings could do the same.

Indeed, definitions in dictionaries are conceived to define compositionally the meaning of target words. Therefore, these are natural candidates for deriving a word embedding of a OOV word by composing the word embeddings of the words in the definition. The hunch is that universal sentence embedders can be used for this purpose. 

Moreover, these definitions have a recurrent structure, which can be definitely used to derive simpler model. Definitions for words $w$ are often organized as a particular sentence which contains the super-type of $w$ and a modifier, which specializes the super-type. For example (Fig. \ref{fig:general_idea}), \emph{cheerlessness} is defined as \emph{a feeling}, which is the super-type, and \emph{of dreary and pessimistic sadness}, which is the modifier. By using this structure, we propose a simpler model for composing meaning.

In the following sections, we propose two models: (1) \Funtional{}, a model that exploit the structure of the definitions to focus on relevant words; and (2) \DefBERTplain{}, a model that utilizes BERT as universal sentence embedder to embed the definition in a single vector. 


\subsection{\Funtional{}: a feed-forward neural network to learn word embedding from definitions} \label{sec:DefNN}

The Definition Neural Network (\Funtional{}) is our first model and has two main components (see Figure \ref{fig:general_idea}). The first component, \emph{DefAnalyzer}, aims to spot the two important words of the definition: the super-type $w_h$ and the main word $w_m$ of the modifier of the super-type. The second component, \emph{DeNN}, is a feed-forward neural network that takes in input the embeddings, $\vec{w}_h$ and $\vec{w}_m$, of the two selected words and produces the embedding for the target word $\vec{w}_{def}$.


To extract the two main words from a given definition, \emph{DefAnalyzer} exploits the recurrent structure of definitions by using their syntactic interpretations. In our study, we use constituency parse trees and correlated rules to extract the super-type $w_h$ and its closest modifier $w_m$. Basically, the simple algorithm is the following. Given a definition $s$, parse the definition $s$ and select the main constituent. If the main constituent contains a semantic head and a modifier, then those are the two target words. In the other case, select the semantic head of the main constituent as the super-type $w_h$ and the semantic head of the first sub-constituent as the relevant modifier $w_m$. For example, the parse tree for the definition of $cherlessness$ in Fig. \ref{fig:general_idea} is the following:
\begin{center}
\resizebox{4cm}{!}{
\Tree [.NP [.NP [.DT a ] [.NN \textbf{feeling} ] ] [.PP [.IN of ] [.NP  [.ADJP [.JJ dreary ] [.CC or ] [.JJ pessimistic ] ] [.NN \textbf{sadness} ] ] ] ] }
\end{center}
In this case the main constituent is the first $\mathrm{NP}$: the selected $w_h$ is the word $\mathrm{feeling}$ which is semantic head of the first $\mathrm{NP}$; $w_m$ is noun  $\mathrm{sadness}$ which is the semantic head of $\mathrm{PP}$. The semantic heads are computed according to a slightly modified version of the semantic heads defined by \citealp{collins-2003-head}.

The second component is \emph{DeNN} that, given the words embbedings $\vec{w}_h$ and $\vec{w}_m$ from the Word2Vec embedding space for respectively $w_h$ and $w_m$ from the definition, their POS tag $pos_h$, $pos_m$ and the target's POS tag $pos_c$ as additional information, outputs the embedding $\vec{w}_{c}$ for the target word $w_c$. The input of \Funtional{} is illustrated in Fig.\ref{fig:general_idea}.
The general equation for \emph{DeNN} is:
\[
    \vec{w}_{c} = \mathbf{DeNN}(\vec{w}_h, \vec{w}_m, pos_h, pos_m, pos_c)
\]

The $\mathbf{DeNN}$ function can be described starting from three simpler subnets: (1)  $\mathbf{FF}_{w}$ processes word embeddings $\vec{w}_h$ and $\vec{w}_m$; (2) $\mathbf{FF}_{p}$ embeds and processes $pos_h$, $pos_m$ and $pos_c$; finally, (3) $\mathbf{FF}$ processes the joint information from the previous steps. 

The equation describing the subnet $\mathbf{FF}_{w}$ that takes as input $\vec{w}_h$ and $\vec{w}_m$ is the following:
\begin{equation}
    \label{equation:sum}
    \vec{s} = \mathbf{FF}_{w}(\vec{w}_h, \vec{w}_m) =  \sigma(\mathbf{W}_{h}\vec{w}_h + \mathbf{W}_{m}\vec{w}_m) 
\end{equation}
where $\mathbf{W}_{h}$ and $\mathbf{W}_{m}$ are dense layer and $\sigma$ is the LeakyReLU activation function. 

The subnet $\mathbf{FF}_{p}$ processes POS tags: $pos_{h}$, $pos_{m}$, $pos_{c}$. Each $pos_i$ for $i \in \{h,m,c\}$ is firstly fed into an embedding layer $\epsilon$ which weights are learned from scratch. The resulting embedding $\epsilon(pos_{i})$ is then fed into a dense layer $\mathbf{W}_{i}$. Hence for each for $i \in \{h,m,c\}$ the output of $\mathbf{FF}_{p}$ is:
\begin{equation}
    \label{equation:embedpos}
    \vec{p}_{i} = \mathbf{FF}_{p}(pos_{h},pos_{m}, pos_{c})[i] = \mathbf{W}_{i}\epsilon(pos_{i})
\end{equation}
The $\vec{s}$ resulting from Equation \ref{equation:sum} and the $\vec{p}_{h}$, $\vec{p}_{m}$, $\vec{p}_{c}$ from the Equation \ref{equation:embedpos} are hence concatenated ($\oplus$):
\[
    \vec{h} = \vec{s} \oplus \vec{p}_{h} \oplus \vec{p}_{m} \oplus \vec{p}_{c}
\]

As final step $\vec{h}$ is fed into a feed-forward subnet $\mathbf{FF}$ composed of the dense layers $\mathbf{W}_{1}$, $\mathbf{W}_{2}$ and $\mathbf{W}_{3}$ as follows:
\begin{equation}
    \label{equation:ff}
    \mathbf{FF}(\vec{h}) = 
    \mathbf{W}_{3}\sigma(\mathbf{W}_{2}(\sigma(\mathbf{W}_{1}\vec{h})))
\end{equation}
Hence the following:
\[
    \vec{w}_{c} = 
    \mathbf{FF}(\mathbf{FF_{w}}(\vec{w}_h, \vec{w}_m),  \mathbf{FF_{p}}(pos_h, pos_m, pos_c))
\]
describes how \emph{DeNN} computes the embedding $\vec{w}_{c}$ for an OOV word having as input $\vec{w}_h$, $\vec{w}_m$, $pos_h$, $pos_m$ from \emph{DefAnalyzer} and $pos_c$. 

For comparative purposes, we defined two additional baseline models: an hyperonym model ($Head$) and an additive model ($Additive$) \cite{mitchell-lapata-2008-vector}. The 
$Head$ model derives the embedding for the OOV word $c$ by using the embedding for its hypernym $h$ in WordNet, that is, $\vec{w}_{c} = \vec{w}_h$. 
The $Additive$ model instead adds the embeddings of the two words in the definition used by \Funtional{}, that is, $\vec{w}_{c} = \vec{w}_h + \vec{w}_m$.

\subsection{\DefBERTplain{}:  Transforming definitions in word embeddings}
\label{sec:DefBERT}
\DefBERTplain{} aims to use BERT's ability to process sentences in order to use directly the definition for $w_c$ in order to produce its embedding $\vec{w}_{c}$. \DefBERT{} and \HeadBERT{} are the approaches followed in exploiting the definition.

\DefBERT{} is the first of these approaches: in this case the definition of $w_c$ is given in input to a pretrained Bert-base model and, as showed in Figure \ref{fig:general_idea}, $\vec{b}_{[CLS]}$, the embedding for the [CLS] token, is taken as sentence embedding in the USE acceptation of BERT. 

\HeadBERT{} is the second approach and in this case is selected $\vec{b}_{head}$, which is contextual emedding of $\vec{w_{h}}$ from the definition. Since BERT's embedding are contextual, $\vec{b}_{head}$ could benefit of the definition being the input sentence. 

For comparative purposes, we also define \OOVBERT{} and \ExampleBERT{}. \OOVBERT{} is used to see if our model outperforms the classical behavior of BERT when it encounters OOV words. Hence, \OOVBERT{} replicates this classical behavior. In this case, BERT is fed with a sample sentence containing the target OOV word, for example \emph{``... melancholy to pastel cheerlessness''} for the target OOV \emph{``cheerlessness''} (see Figure \ref{fig:general_idea}). Then, the word is divided in word pieces. To obtain the embedding for the target word, we sum up vectors of these word pieces. \ExampleBERT{} instead is used to determine if definitions are really useful for modeling meaning of the head word. \ExampleBERT{} is similar to \HeadBERT{} but the input is different. \ExampleBERT{} has a random sentence which contains the head word. Hence,  comparing  \HeadBERT{} with \ExampleBERT{} gives intuition if the head in definition really absorbs its meaning.       

\section{Experiments}
Experiments want to investigate three issues: (1) if word embeddings  obtained with \Funtional{} are reasonably better than baseline compositional functions to obtain embeddings as well as those obtained with an untrained version of BERT; (2) if similarity measures over WordNet are correlated with spaces of word embeddings; (3) finally, if word embeddings for Out-of-Vocabulary words obtained are good word representations in terms of their correlation with similarity measures on WordNet. Clearly, issue (2) is necessary to investigate issue (3) and we spend time to analyze issue (2) as the correlation between WordNet measures and word embeddings is a highly debated problem \cite{LASTRADIAZ2019645}.

The rest of the section is organized as follows. Section \ref{sec:exp_setup} introduces the general settings of our experiments. Section \ref{sec:results} presents results and it is organized in three subsections, which address the above three issues. If needed, these subsections introduce additional settings for the experiments.

\subsection{Experimental set-up}
\label{sec:exp_setup}
Our experiments are primarily defined around WordNet \cite{wordnet-resource}. WordNet is the source of word definitions, which are needed for \Funtional{} and for \DefBERTplain{}. WordNet is used to collect testing sets of word pairs of similar and dissimilar words. Finally, similarity measures over WordNet are used to rank pairs according to the similarity between words. These latter rankings are used to see if similarities derived with \Funtional{}'s and \DefBERTplain{}'s word embeddings for OOV words correlate with a standard notion of similarity between two words.   

In our study, in-vocabulary (IV) and OOV words ($IV_{w2v}$, $OOV_{w2v}$, $IV_{BERT}$ and $OOV_{BERT}$) are defined according to a pre-trained word embedding matrix $W_{w2v}$ and $W_{BERT}$. 
$W_{w2V}$ is the Word2Vec's embedding space \cite{mikolov2013efficient}  pre-trained on part of Google News dataset (about 100 billion words) and $W_{BERT}$ is the BERT's word embedding space \cite{devlin-etal-2019-bert} trained on lower-cased English text from BooksCorpus (800M words) \cite{DBLP:journals/corr/ZhuKZSUTF15} and English Wikipedia (2,500M words) as described in \citeauthor{devlin-etal-2019-bert}. 
Then, $IV_{w2v}$ and $IV_{BERT}$ words are words in WordNet that are in the target embedding matrix and $OOV_{w2v}$ and $OOV_{BERT}$ are words in WordNet that are not in the target embedding matrix. These OOV words are interesting since, in principle, their meaning is known in WordNet but their embedding is not available. Then, \Funtional{} as well as \DefBERTplain{} can be definitely utilized. In selecting $IV_{BERT}$ and $OOV_{BERT}$, there is an additional limitation: in order to apply \DefBERTplain{}, usage examples are needed. Then, $IV_{BERT}$ and $OOV_{BERT}$ are words that have an usage example in WordNet. 

We prepared two different sets of datasets for \emph{directly} and \emph{indirectly} investigating \Funtional{} and \DefBERTplain{}. 

In the \emph{direct} investigation, \Funtional{} and \DefBERTplain{} are tested to verify their ability to produce vectors for IV words. Methods are compared on the distribution (mean and standard deviation) of cosine similarity between the embedding of words and the embedding produced by using their definitions. We then have selected: 1) $Train_{v2w}$ with 33404 words and $Test_{v2w}$ with 8336 words as subsets of $IV_{w2v}$; 2) $Test_{BERT}$ with 3218 words as subset of $IV_{BERT}$.  $Train_{v2w}$ is also used to train \Funtional{}.

In the \emph{indirect} investigation,  \Funtional{} and \DefBERTplain{} are tested to assess their ability to produce embeddings for OOV that may replicate some similarity measure between words in pairs. 
We selected three similarity measures defined over WordNet: $path$ \citep{path_aricle}, $wup$ \citep{wu-palmer-1994-verb} and $res$ \citep{resnik1995using}. 
Then, we collected two sets of pairs of words $Pairs_{w2v}$ and $Pairs_{BERT}$. Word pairs $(w_1,w_2)$ in $Pairs_{w2v}$ are selected as follows: (1) $w_1$ is in $OOV_{w2v}$; (2) $w_2$ is in $IV_{w2v}$ is either a random sister word of $w_1$ in 50\% of the cases or a random word in the other 50\% of cases. Word pairs $(w_1,w_2)$ in $Pairs_{BERT}$ are obtained similarly. $Pairs_{w2v}$ contains around 4,500 word pairs and $Pairs_{BERT}$ contains 3500 word pairs. To correctly apply Spearman's correlation between our systems and the expected rank on the list of pairs induced by a similarity measure, we divided $Pairs_{w2v}$ in 600 lists of 7 pairs and $Pairs_{BERT}$ in 450 lists. $Pairs_{w2v \cap BERT}$ contains 450 pairs divided into 60 lists. 
Pairs in the list are selected to have 7 clearly different values of the selected similarity ($path$, $wup$ and $res$) between the two words. The final Spearman's correlation is a distribution of correlation over these lists.  

The last datasets here defined are used to investigate the second issue addressed at the beginning of this section: it is necessary to determine if measures over WordNet are correlated with spaces of word embedding.
The investigated words embeddings are Word2Vec, FastText, BERT.
Similarly to $IV_{w2v}$ and $IV_{BERT}$, $IV_{fasttext}$ is a set of word in WordNet that are in the $W_{fasttext}$ target embedding matrix of FastText.
$Pairs_{IV_{w2v}}$, $Pairs_{IV_{BERT}}$ and $Pairs_{IV_{fasttext}}$ are the built dataset and each of them is composed of pairs $(w_1,w_2)$ of words from the given $IV$, where $w_2$ is either a random sister word of $w_1$ in 50\% of the cases or a random word in the other 50\% of cases. 
This definition follows the same approach used in defying  $Pairs_{w2v}$ and $Pairs_{BERT}$.
$Pairs_{IV_{w2v}}$, $Pairs_{IV_{BERT}}$ and $Pairs_{IV_{fasttext}}$ contain respectively about 14,000, 560 and 14,000 pairs. These are then divided into smaller lists of 7 pairs where Spearman's coefficient is computed.

To comparatively investigate our \Funtional{} and \DefBERTplain{}, we used FastText \cite{bojanowski2016enriching} as realized in \citet{grave2018learning} along with: (1) Additive and Head defined in Section \ref{sec:DefNN}; (2) \OOVBERT{} and \ExampleBERT{} defined in Section \ref{sec:DefBERT}.  
FastText defines embeddings unknown words $c$ by combining embeddings of 3-grams, for example, the embedding for the OOV word \emph{cheerlessness} is represented as the vector $\vec{f_c} = \vec{che} + \vec{hee} + ... + \vec{ess}$.

As final experimental setting, definitions are parsed using Stanford’s CoreNLP probabilistic context-free grammar parser \cite{manning-EtAl:2014:P14-5}. NLTK  \citep{journals/corr/cs-CL-0205028} is used to access WordNet and compute similarity measures over it.

\subsection{Results and discussion}
\label{sec:results}

For clarity, this section is organized around the three issues we aim to investigate: the ability of proposed methods to build embeddings of words starting from dictionary definitions (Sec.~\ref{sec:similarity}); the debated relation between similarity over word embeddings and similarity in WordNet (Sec.~\ref{sec:distributionalVSwordnet}); and, finally, the ability of the proposed methods to produce embeddings for OOV words (Sec.~\ref{sec:DiscoveringOOVembedding}).  


\subsubsection{Word Embeddings from Dictionary Definitions}
\label{sec:similarity}

The first issue to investigate is whether our methods produce word embeddings from dictionary definitions that are similar with respect to word embeddings directly discovered. We then studied the cosine similarity between the two kinds of embeddings, for example, between the embedding of \emph{cheerlessness} and the embedding of the definition \emph{a feeling of .... sadness}. For the diffent methods, the comparison is on their own space, that is, $sim(\vec{w}_c,\vec{w}_{def})$ for \Funtional{} and $sim(\vec{b}_c,\vec{b}_{[CLS]})$ or $sim(\vec{b}_c,\vec{b}_{head})$ for \DefBERT{} and \HeadBERT{}, respectively (see Fig. \ref{fig:general_idea}). Experiments are conducted on In-Vocabulary words for both spaces by using the $Test_{w2v}$, $Test_{BERT}$ and $Test_{w2v \cap BERT}$ datasets.

\begin{table}[h]
\resizebox{\linewidth}{!}{
\begin{tabular}{|l|l c c|} 
 \hline
                &               & \textbf{nouns} & \textbf{verbs}\\
 \emph{Dataset} &  \emph{Model} & \emph{sim} & \emph{sim}\\ 
 \hline\hline
    \multirow{3}{*}{$Test_{w2v}$}
    & Additive & $0.28 (\pm 0.16)^\circ$ & $0.30 (\pm 0.19)^\circ$\\
    & Head & $0.27 (\pm 0.20)^\star$ & $0.30 (\pm 0.26)^\star$\\
    & \Funtional{} & $\mathbf{0.46 (\pm 0.14)}^{\circ\star}$  & $\mathbf{0.48 (\pm 0.13)}^{\circ\star}$ \\
    \hline
    \hline
    \multirow{3}{*}{$Test_{BERT}$}
    & \HeadBERT{} & $\mathbf{0.46(\pm 0.13)}^{\dag\ddag}$ & $\mathbf{0.41(\pm 0.14)}^{\dag\ddag}$\\
    & \DefBERT{} & $0.32 (\pm 0.08)^\dag$ & $0.30 (\pm 0.09)^\dag$\\
    & \ExampleBERT{} & $0.41 (\pm 0.12)^\ddag$ & $0.39 (\pm 0.12)^\ddag$\\
    \hline
    \hline
    \multirow{3}{*}{$Test_{w2v \cap BERT}$}
    & \HeadBERT{} & $\mathbf{0.48(\pm 0.12)}^{\dag\triangle}$ & $0.43(\pm 0.15)^{\dag\triangle}$\\
    & \DefBERT{} & $0.30 (\pm 0.09)^{\dag\diamond}$  & $0.28 (\pm 0.09)^{\dag\diamond}$ \\
    & \Funtional{} & $0.37 (\pm 0.12)^{\triangle\diamond}$ & $\mathbf{0.49 (\pm 0.12)}^{\triangle\diamond}$\\
    \hline
 \hline
\end{tabular}
}
\caption{Cosine similarity between word embeddings and embeddings of their definitions. The marking signs $\star$, $\circ$, $\dag$, $\ddag$ and $\diamond$ indicate pairs of models results for which the higher result is statistically significant better than the other (with a 95\% confidence level) according to the one sided Wilcoxon signed-rank test.}
\label{table:Table 1}
\end{table}

Definitions seem to be better sources of word embeddings instead of baseline methods and other solutions. In fact, both \Funtional{} and \HeadBERT{} outperform different methods in their respective tests for both nouns and verbs (see Table \ref{table:Table 1}). For nouns, \Funtional{} has an average cosine similarity of $0.46(\pm0.14)$, which is well above that of Additive ($0.28(\pm16)$) and Head ($0.27(\pm20)$). In the same syntactic category, \HeadBERT{} outperforms \ExampleBERT{}, $0.46(\pm0.13)$ vs. $0.41(\pm0.12)$. For verbs, \Funtional{} has an average cosine similarity of $0.48(\pm 0.13)$, which is well above the Additive and the Head. In the same category, \HeadBERT{} slightly outperforms \ExampleBERT{}. Finally, in the common test, that is, $Test_{w2v \cap BERT}$, definition based models outperform simpler models. \HeadBERT{} has a better similarity for nouns and \Funtional{} has a better similarity for verbs. 

For BERT, the embedding emerging related to the token [CLS] does not seem to represent the good token where to take semantics of the sentence in terms of a real composition of the meaning of component words. \DefBERT{} performs poorly with respect to \HeadBERT{} and also with respect to \ExampleBERT{} in both syntactic categories for $Test_{BERT}$ (see Table \ref{table:Table 1}). This is confirmed in the restricted set $Test_{w2v \cap BERT}$. Therefore, even if the embedding in token [CLS] is often used as universal sentence embedding for classification purposes \cite{devlin-etal-2019-bert, adhikari2019docbert, jiang-de-marneffe-2019-evaluating}, it may not to contain packed meaning whereas it may contain other kinds of information regarding the sentence.


\subsubsection{Word Embedding Spaces and WordNet}
\label{sec:distributionalVSwordnet}

WordNet and it's correlated similarly metrics can be an interesting opportunity to extract testsets for assessing whether our methods can be used to derive embeddings of OOV words.
However, it is a strongly debated question whether similarities in WordNet are correlated with similarities over word embeddings \cite{LASTRADIAZ2019645}.

The aim of this section is twofold. Firstly, it aims to investigate if this relation can be established on the word embedding spaces we are using. Secondly, it aims to validate and select plausible similarity measures over WordNet, which can then be used to investigate the behavior of embeddings for OOV words. For both experimental sessions, we used the datasets  $Pairs_{IV_{w2v}}$, $Pairs_{IV_{BERT}}$ and $Pairs_{IV_{fasttext}}$ definde in Section \ref{sec:exp_setup}.

\begin{table}[h!]
\resizebox{\linewidth}{!}{
\begin{tabular}{|c c c c|} 
 \hline
 \emph{Model} & \emph{Dataset} & \emph{Category} & \emph{AUC value}\\ 
 \hline
    \hline
    \multirow{2}{*}{Word2Vec} 
    & \multirow{2}{*}{$Pairs_{IV w2v}$}
    & verbs & $0.64$\\
    & & nouns & $0.79$ \\
    \hline
    \multirow{2}{*}{FastText}
    & \multirow{2}{*}{$Pairs_{IV fasttext}$}
    & verbs & $0.63$\\
    & &nouns & $0.82$ \\
    \hline
    \multirow{2}{*}{BERT}
    & \multirow{2}{*}{$Pairs_{IV BERT}$}
    & verbs & $0.73$\\
    & &nouns & $0.68$ \\
    \hline
\end{tabular}
}
\caption{AUC value in classifying words sister terms}
\label{table:Table AUC in vocabulary sister terms}
\end{table}
For the first aim, we investigated whether similarities derived on a particular word embedding spaces can be used to divide positive and negative pairs in the respective sets of pairs. Then, given a word embedding space, we ranked pairs according to computed similarities and we computed the Area under the ROC built on sensitivity and specificity.


Results show that there is a correlation between "being siblings" and the three word embedding spaces, $w2v$, $BERT$ and $fasttext$ (Table \ref{table:Table AUC in vocabulary sister terms}). All the AROCs are well above the threshold of 0.5 and close or above the value of 0.7, which indicates a good correlation.

\begin{table}[h!]
\resizebox{\linewidth}{!}{
\begin{tabular}{|c c c c|} 
 \hline
 \emph{Model} & \emph{Dataset} & \emph{Measure} & \emph{Spearman}\\ 
 \hline
    \hline
    \multirow{3}{*}{Word2Vec}
    & \multirow{3}{*}{$Pairs_{IV w2v}$}
    & $path$ & $0.25 (\pm 0.39)$\\
    & & $wup$ & $0.25	(\pm 0.38)$ \\
    & & $res$ & $0.50	(\pm 0.31)$ \\
    \hline
    \multirow{3}{*}{FastText}
    & \multirow{3}{*}{$Pairs_{IV fasttext}$}
    & $path$ & $0.31 (\pm 0.38)$\\
    & & $wup$ & $0.40 (\pm 0.35)$ \\
    & & $res$ & $0.52 (\pm 0.29)$ \\
    \hline
    \multirow{3}{*}{BERT}
    & \multirow{3}{*}{$Pairs_{IV BERT}$}
    & $path$ & $0.08(\pm 0.40)$\\
    & & $wup$ & $0.29 (\pm 0.39)$ \\
    & & $res$ & $0.28 (\pm 0.38)$ \\
    \hline
\end{tabular}
}
\caption{Average Spearman Coefficient measuring correlation on cosine similarity among embedding and similarity over WordNet taxonomy}
\label{table:Table in voc spearman}
\end{table}

\begin{table*}[h!]
\begin{center}
\begin{tabular}{|l|l c c c|} 
 \hline
 \emph{Dataset}  & \emph{Model} & \emph{Corr(path)} & \emph{Corr(wup)} & \emph{Corr(res)}\\ 
 \hline\hline
    \multirow{4}{*}{$Pairs_{w2v}$}
    &Additive 				& $\mathbf{0.43 (\pm 0.33)}$ & $0.54 (\pm 0.29)$& $0.45 (\pm 0.33)^\circ$ \\
    &Head 					& $0.41 (\pm 0.34)$ & $\mathbf{0.57 (\pm 0.33)}$& $0.48 (\pm 0.36)^\star$\\
    &FastText 				& $0.29 (\pm  0.37)$& $0.42 (\pm 0.36)^{\diamond}$& $0.34 (\pm 0.37)^\diamond$\\
    &\Funtional{} 			& $0.30 (\pm  0.34)$& $0.56 (\pm 0.30)^{\diamond}$& $\mathbf{0.51 (\pm 0.31)}^{\circ\star\diamond}$\\
    \hline\hline                                                           
    \multirow{4}{*}{$Pairs_{BERT}$}
    &\HeadBERT{} 			& $\mathbf{0.27 (\pm 0.36)}^{\ddag\bullet}$ & $\mathbf{0.33	(\pm0.37)}^{\dag\ddag\bullet}$ & $\mathbf{0.30	(\pm 0.36)^{\dag\ddag\bullet}}$ \\           
    &\DefBERT{} 			& $0.26 (\pm 0.36)$ & $0.17	(\pm0.37)\dag$ & $0.11	(\pm 0.39)^\dag$ \\
    &\ExampleBERT{} 			& $0.15 (\pm 0.41)^\ddag$ & $0.25	(\pm0.38)^\ddag$ & $0.19	(\pm 0.40)^\ddag$ \\
    &\OOVBERT{} 			& $0.09 (\pm 0.37)^\bullet$ & $0.19	(\pm0.37)^\bullet$ & $0.23	(\pm 0.38)^\bullet$ \\
    \hline\hline                                                           
    \multirow{4}{*}{$Pairs_{w2v \cap BERT}$}                              
    &\HeadBERT{} 			& $0.33 (\pm 0.32)^{\triangle\bullet}$ & $0.27	(\pm0.37)^{\triangle\bullet}$ & $0.23	(\pm 0.39)^{\triangle\bullet}$ \\
    &\Funtional{} 			& $\mathbf{0.42 (\pm 0.31)}^{\triangle}$ & $\mathbf{0.44 (\pm0.32)}^{\triangle} $ & $\mathbf{0.39	(\pm 0.34)}^{\triangle\diamond}$\\
    &FastText 				& $0.38 (\pm 0.38)$ & $0.37	(\pm 0.34)$ & $0.30	(\pm 0.35)^\diamond$ \\
    &\OOVBERT{} 			& $0.02 (\pm 0.41)^{\bullet}$ & $0.10	(\pm 0.39)^{\bullet}$ & $0.15	(\pm 0.39)^{\bullet}$\\
    \hline
 \hline
\end{tabular}
\end{center}
\caption{Average Spearman coefficient from the \emph{indirect} investigation. The marking signs $\star$, $\circ$, $\bullet$, $\dag$, $\ddag$, $\triangle$ and $\diamond$ indicate pairs of models results for which the higher result is statistically significant better than the other (with a 95\% confidence level) according to the one sided Wilcoxon signed-rank test.}
\label{table:Table microlistOOV}
\end{table*}

For the second aim, we investigated WordNet Similarity metrics in order to find interesting metrics to experiment with our definition-oriented methodologies. In fact, the binary task of being or not being siblings in WordNet may not capture real nuances of similarity as word embeddings are capturing. Sibling words may be very similar or less similar. For example, \emph{cheerlessness} and \emph{depression} (see Figure \ref{fig:general_idea}) are sibling words and are definitely similar. On the contrary, \emph{house} and \emph{architecture} are sibling words but are less similar with respect to the previous pair of words. In WordNet, this difference in similarity is captured by using many different metrics.  


We investigated three different WordNet similarity measures: $path$ \citep{path_aricle}, $wup$ \citep{wu-palmer-1994-verb} and $res$ \cite{resnik1995using}. The measure $path$ uses the length of the path connecting two synsets over the WordNet taxonomy.
The measure $wup$ is still based on the length of path between the synsets related to the two words and takes into account the number of edges from synsets to their Least Common Subsumer (LCS) and the number of links from the LCS up to the root of the taxonomy. Finally, the measure $res$ belongs to another family of measures as it is based on the Information Content. In $res$, the similarity between synsets of the related words is a  function of the Information Content of their LCS. In this case, a more informative LCS (a rare as well as a specific concept) indicates that the hyponym concepts are more similar.

The best correlated WordNet measure is $res$. In fact, it is highly correlated for two spaces out of three, Word2Vec and FastText, and it is on par with $wup$ in the BERT space (see \ref{table:Table in voc spearman}). The average Spearman's correlation between the word embedding spaces of Word2Vec and $res$ is $0.50(\pm0.31)$, which is well above $path$ and $wup$. The same happens for the space FastText where the correlation is  $0.52(\pm0.29)$. 

As a final consideration, for our purposes, word embedding spaces are correlated and the best measure that captures this correlation is $res$.


\subsubsection{Testing over Out-Of-Vocabulary Words}
\label{sec:DiscoveringOOVembedding}


The final analysis is on real OOV words for Word2Vec and for BERT. These last experiments are carried out by considering the positive relation between WordNet similarity measures and the word embedding spaces.

Using definitions for deriving word embeddings for OOV words seems to be the good solution compared to alternative available approaches. 

In its space, \Funtional{} achieves very important results for the correlation with the two WordNet similarity measures $wup$ and $res$ (see Table \ref{table:Table microlistOOV}). In both cases, it outperforms FastText, which is a standard approach for deriving word embeddings fo OOV words ($0.51\pm0.31$ vs. $0.34\pm0.37$ for $res$ and $0.56\pm0.30$ vs. $0.42\pm0.36$ for $wup$). Moreover, \Funtional{} outperforms Head, a baseline method based on WordNet, and Additive, the simplest model to use WordNet definitions.

The same happens for \HeadBERT{} in its space (see Table \ref{table:Table microlistOOV}). \HeadBERT{} significantly outperforms \OOVBERT{}, showing that \HeadBERT{} is a better model to treat OOV with respect to that already included in BERT. Results on \HeadBERT{} confirms that the output related to the token representing the head carries better information than the output related to the token [CLS]. Moreover, the definition has is a positive effect on shaping the word embedding of the head word towards the defined word. In fact, \HeadBERT{} and \ExampleBERT{} are applied on the same head word and \HeadBERT{} transforms better the meaning than \ExampleBERT{}, which is applied to a random sentence containing the head word. Indeed, also for BERT, definitions are important in determining embeddings of OOV words.

The final comparison is between \Funtional{} and \HeadBERT{} and it is done on the small dataset $Pairs_{w2n\cap BERT}$. 
\Funtional{} outperforms \HeadBERT{} for all the three WordNet measures (see Table \ref{table:Table microlistOOV}). These results show that the simpler is the better in using definitions for OOV words.

\section{Conclusions and Future Work}


Building word embedding for rare out-of-vocabulary words is essential in natural language processing systems based on neural networks. In this paper, we proposed to use definitions in dictionaries to solve this problem. Our results show that this can be a viable solution to retrieve word embedding for OOV rare words, which work better than existing methods and baseline systems. 

Moreover, the use of dictionary definitions in word embedding may open also another possible line of research: a different semantic probe for universal sentence embedders (USEs). Indeed, 
definitions offer a definitely interesting equivalence between sentences and words. Hence, unlike existing semantic probes, this approach can unveil if USEs are really changing compositionally the meaning of sentences or are just aggregating pieces of sentences in a single representation. 



\newpage
\bibliography{anthology,custom}
\bibliographystyle{acl_natbib}

\end{document}